\documentclass[letterpaper, 10 pt, conference]{ieeeconf}  

\IEEEoverridecommandlockouts                              

\usepackage{mathptmx} 
\usepackage{multirow}
\usepackage{mathtools}
\usepackage{amsmath} 
\usepackage{amssymb}  
\usepackage{nccmath}
\usepackage{graphicx}
\usepackage{gensymb}
\usepackage{commath}
\usepackage{float}
\usepackage[table,xcdraw]{xcolor}
\usepackage{makecell}
\usepackage{algorithm}
\usepackage{algpseudocode}
\usepackage{cite}
\usepackage[hidelinks]{hyperref}
\usepackage{nicematrix}
\usepackage{url}
\usepackage{bm}
\usepackage{hhline}
\usepackage{booktabs}
\usepackage{float}
\usepackage{comment}
\usepackage{subfloat}
\usepackage{mhchem}

\usepackage[caption=false,font=footnotesize]{subfig}

\title{\LARGE \bf
Proprioceptive External Torque Learning for Floating Base Robot and its Applications to Humanoid Locomotion*
}

\author{Daegyu Lim$^{1}$, Myeong-Ju Kim$^{1}$, Junhyeok Cha$^{1}$, Donghyeon Kim$^{1}$ and Jaeheung Park$^{1,2}$
\thanks{*This work was supported by the National Research Foundation of Korea (NRF) grant funded by the Korea government (MSIT) (No. 2021R1A2C3005914).}
\thanks{$^{1}$Daegyu Lim, Myeong-Ju Kim, Junhyeok Cha, Donghyeon Kim and Jaeheung Park are with the Department of Intelligence and Information, Seoul National University, Seoul, Republic of Korea. {\tt\footnotesize \{dgyo3784, myeong-ju, threeman1, kdh0429, park73\}@snu.ac.kr}}%
\thanks{$^{2}$Jaeheung Park is also with the Advanced Institutes of Convergence Technology, Republic of Korea, and with ASRI, RICS, Seoul National University, Republic of Korea.}%
}

\begin{document}

\maketitle
\thispagestyle{empty}
\pagestyle{empty}

\begin{abstract} 
The estimation of external joint torque and contact wrench is essential for achieving stable locomotion of humanoids and safety-oriented robots. Although the contact wrench on the foot of humanoids can be measured using a force-torque sensor (FTS), FTS increases the cost, inertia, complexity, and failure possibility of the system. This paper introduces a method for learning external joint torque solely using proprioceptive sensors (encoders and IMUs) for a floating base robot. For learning, the GRU network is used and random walking data is collected. Real robot experiments demonstrate that the network can estimate the external torque and contact wrench with significantly smaller errors compared to the model-based method, momentum observer (MOB) with friction modeling. The study also validates that the estimated contact wrench can be utilized for zero moment point (ZMP) feedback control, enabling stable walking. Moreover, even when the robot's feet and the inertia of the upper body are changed, the trained network shows consistent performance with a model-based calibration. This result demonstrates the possibility of removing FTS on the robot, which reduces the disadvantages of hardware sensors. The summary video is available
at \url{https://youtu.be/gT1D4tOiKpo}.
\end{abstract}

\section{Introduction}
\label{Section/Introduction}
Maintaining balance while walking in the presence of disturbances is the most fundamental and crucial ability for humanoids.
Among the many technologies, contact wrench (linear force, and moment) control significantly contributes to realizing robust balancing control against disturbances.
For humanoids with high gear ratio actuators\cite{stasse2017talos, kaneko2019humanoid}, external wrench or external joint torque sensing ability is crucial to control the contact wrench precisely using the external wrench feedback because the large friction torque and the amplified rotor inertia by the high reduction gear deteriorate the feedforward joint torque control performance. Therefore, these human-size humanoids are equipped with joint torque sensors (JTS) in each joint or force-torque sensors (FTS) on the feet. However, additional sensors increase not only the cost of the robot but also the inertia, complexity, and failure possibility of the system. 
On the other hand, some bipedal robots such as Cassie\cite{abate2018mechanical} or MIT humanoid \cite{chignoli2021humanoid} are built with low gear reduction ratio actuators to enhance the torque-transparency. The open loop torque control strategy is used for these robots by assuming that the commanded torque can be realized without any force-torque sensor feedback. However, even for robots with quasi-direct drive motors, proprioceptive external torque estimation is necessary for more accurate contact wrench feedback control and collision handling.

Studies on the external disturbance estimation for humanoids using the dynamics have been conducted \cite{benallegue2018model, flacco2016residual, vorndamme2017collision, lee2018contact}. However, all the methods still use FTSs for contact wrench estimation and these model-based methods suffer from dynamics modeling error and other dynamics which is not considered in the model such as joint friction or elasticity.
In \cite{lee2015sensorless}, the joint friction is modeled and reflected on the momentum residuals calculation, but finding a proper friction model is hardware-dependent, and it is difficult to fit the model accurately.

To resolve the disadvantages of the model-based external torque estimation, we previously proposed the momentum observer (MOB) based model-uncertainty learning method \cite{lim2021momentum}. Using a long short-term memory (LSTM) network, a kind of recurrent neural network (RNN), the uncertainty torque including joint friction and modeling error is learned during the free motion. However, the previous method is applied to a 2-DoF fixed-based manipulator, and the estimated pure external torque is only used for the collision detection task. A similar approach to our previous work is applied for the FTS attached to the end-effector of the collaborative robot, KUKA, in \cite{dine2018force}. Dine et al. \cite{dine2018force} proposed an observer based on a recurrent neural network (RNNOB) to learn the non-contact force measurement of the FTS and offset the disturbance measurement. For learning the non-contact force, only encoders and inertial measurement unit (IMU) on the end-effector are used, but FTS is still required after the learning. The data-driven torque estimation method is also applied to the surgical robot (da Vinci) \cite{yilmaz2020neural, tran2020deep}. In \cite{yilmaz2020neural}, the joint torque signal for each joint is trained on a small multi-layer perceptron (MLP) with joint position and velocity during free motion. On the other hand, 3D linear force learning is proposed in \cite{tran2020deep} using all the joint torques and joint velocities. However, there is a lack of research focusing on external torque learning for the floating-based robot, especially for humanoids.

Hwangbo et al. \cite{hwangbo2019learning} introduced the actuator network to resolve the sim2real gap when the quadrupedal robot, ANYMAL, is trained to walk in simulation using deep reinforcement learning. The actuator network infers a net joint torque from the proprioceptive input features (history of joint position error and joint velocity). This work shows that the net joint torque can be learned only with the history of the proprioceptive information for each joint. 

In this paper, we propose the external torque learning method for humanoids using only proprioceptive sensors (encoders for each joint and one IMU for the base), and the estimated external torque is used as a feedback signal in the humanoid walking controller. Our contributions can be summarized as follows. First, to the best of our knowledge, it is the first research to learn the external torque using only the proprioceptive sensing for the floating base humanoid robot, and it is validated through various experiments. Second, it is demonstrated that the learned external torque is accurate enough to be used for the zero moment point (ZMP) feedback control of humanoids resulting in stable walking performance. Third, the consistent performance of the external torque estimation for the modification of the robot foot link and upper body mass is validated using real robot walking experiments with model-based calibration. Therefore, this work shows the possibility of substituting the FTS with a neural network, which reduces the cost, inertia, complexity, and the possibility of a sensor failure of the robot system.

\section{Preliminaries}
\label{Section/Preliminaries}

\subsection{Rigid Body Dynamics of Floating Base Robot}
\label{Subsection/Preliminaries/Rigid Body Dynamics of Floaing Base Robot}
A humanoid can be described as a floating base multi-body system that consists of $n+1$ rigid bodies and $n$ joints. When the robot is a floating base robot, the dynamics of the robot can be represented by connecting six virtual joints to the base frame. Thereby, the fundamental rigid body dynamics of an $n+6$ degrees of freedom (DoF) with six virtual joints is as follows
\begin{align}
\label{equation/dynamics equation}
    \mathbf{M}(\mathbf{q}_{v})\ddot{\mathbf{q}}_{v}+\mathbf{C}(\mathbf{q}_{v},\dot{\mathbf{q}}_{v})\dot{\mathbf{q}}_{v} + \mathbf{g}(\mathbf{q}_{v}) & = \bm{\tau}_{v}+\bm{\tau}_{f}+\bm{\tau}_{e}, \\
\label{equation/external torque ft}
    \bm{\tau}_{e} &= \sum_{i=1}^{k} \mathbf{J}^{T}_{c,i}\mathbf{F}_{e,i},
\end{align}
where $ \mathbf{M}(\mathbf{q}_{v}), \mathbf{C}(\mathbf{q}_{v},\dot{\mathbf{q}}_{v})\in \mathbb{R}^{(n+6)\times (n+6)}$, and $\mathbf{g}(\mathbf{q}_{v})\in \mathbb{R}^{n+6}$ are the inertia matrix, the Coriolis and centrifugal matrix, and the gravity vector, respectively. $\mathbf{q}_v, \dot{\mathbf{q}}_{v}, \ddot{\mathbf{q}}_{v}\in \mathbb{R}^{n+6}$ are the generalized position, velocity, and acceleration vectors including virtual joints, respectively. $\bm{\tau}_{v}\in \mathbb{R}^{n+6}$ is the control torque, $\bm{\tau}_{f}\in \mathbb{R}^{n+6}$ is the friction torque, and $\bm{\tau}_{e}\in \mathbb{R}^{n+6}$ is the external torque. $k$, $\mathbf{J}_{c,i}\in \mathbb{R}^{6\times (n+6)}$, and $\mathbf{F}_{e,i}\in \mathbb{R}^{6}$ are the total number of contacts, the $i$-th contact Jacobian matrix, and the $i$-th external wrench, respectively. The generalized coordinate vector consists of 6 virtual joints, and $n$ motor angles: $\mathbf{q}_{v} = [\mathbf{x}_{fb}^{T} \ \mathbf{R}_{fb}^{T} \ \mathbf{q}^{T}]^{T}$, $\dot{\mathbf{q}}_{v} = [\mathbf{v}_{fb}^{T} \ \bm{\omega}_{fb}^{T} \ \dot{\mathbf{q}}^{T}]^{T}$, $\ddot{\mathbf{q}}_{v} = [\dot{\mathbf{v}}_{fb}^{T} \ \dot{\bm{\omega}}_{fb}^{T} \ {\mathbf{\ddot{q}}}^{T}]^{T}$. $\mathbf{x}_{fb}^{T} \in \mathbb{R}^{3}$ and $\mathbf{R}_{fb}^{T}\in SO(3)$ are the position and the orientation of the floating base. $\mathbf{v}_{fb}\in \mathbb{R}^{3}$ and $\bm{\omega}_{fb}\in \mathbb{R}^{3}$ are the linear and angular velocity of the floating base. The input torque has zero elements for the virtual joints, and motor torque $\bm{\tau}_{m}\in \mathbb{R}^{n}$ for the actuating joints as $\bm{\tau}_{v} = [\mathbf{0}^{T} \  \bm{\tau}_{m}^{T}]^{T}$. The friction torque has only joint frictions $\bm{\tau}_{f,j} \in \mathbb{R}^{n}$ as $\bm{\tau}_{f} = [\mathbf{0}^{T} \  \bm{\tau}_{f,j}^{T}]^{T}$.

\subsection{External Torque Estimation using Proprioceptive Sensors}
\label{Subsection/Preliminaries/External Torque Estimation using Proprioceptive Sensors}
The external torque on the robot, $\bm{\tau}_{e}$, can be calculated by measuring the contact wrench $\mathbf{F}_{e}$ using FTS as described in (\ref{equation/external torque ft}). However, if the FTS is not available, the external torque should be estimated from only proprioceptive sensors including the joint encoder, motor current sensor, and IMU. Equation (\ref{equation/dynamics equation}) can be rearranged for the external torque as,
\begin{equation}
\label{equation/external torque}
    \bm{\tau}_{e} = \mathbf{M}(\mathbf{q}_{v})\ddot{\mathbf{q}}_{v}+\mathbf{C}(\mathbf{q}_{v},\dot{\mathbf{q}}_{v})\dot{\mathbf{q}}_{v} + \mathbf{g}(\mathbf{q}_{v}) - \bm{\tau}_{v} - \bm{\tau}_{f}.
\end{equation}
Assuming that the joint friction torque is a function of joint position and velocity $\bm{\tau}_{f,j}(\mathbf{q}, \dot{\mathbf{q}})$, $\bm{\tau}_{e}$ can be expressed in the function of $\mathbf{q}_{v}$, $\dot{\mathbf{q}}_{v}$, $\ddot{\mathbf{q}}_{v}$ and $\bm{\tau}_{m}$ as $\bm{\tau}_{e} = f(\mathbf{q}_{v}, \dot{\mathbf{q}}_{v}, \ddot{\mathbf{q}}_{v}, \bm{\tau}_{m})$.

\subsection{Momentum Observer} 
\label{Subsection/Preliminaries/Momentum Observer}
Joint acceleration $\ddot{\mathbf{q}}$ in (\ref{equation/external torque}) is highly noisy because it is usually measured by the numerical derivative. In order to avoid using noisy signal $\ddot{\mathbf{q}}$ for the calculation of $\bm{\tau}_{e}$, a conventional model-based method, MOB was introduced in \cite{de2003actuator}. The generalized momentum $\mathbf{p}$ of the robot is introduced as
\begin{equation}
\label{equation/generalized momentum}
    \mathbf{p} = \mathbf{M}(\mathbf{q}_{v})\dot{\mathbf{q}}_{v}.
\end{equation}
The derivative of $\mathbf{p}$ can be represented by exploiting the dynamics of the robot in (\ref{equation/dynamics equation}) and the well-known relation $\dot{\mathbf{M}} = \mathbf{C} + \mathbf{C}^{T} $ as
\begin{align}
        \label{derivative of generalized momentum}
        {\dot{\mathbf{p}}} &=  \bm{\tau}_{v} + \bm{\tau}_{f} + \bm{\tau}_{e} +\mathbf{C}^{T}(\mathbf{q}_{v},\dot{\mathbf{q}}_{v})\dot{\mathbf{q}}_{v}-\mathbf{g}_{v}(\mathbf{q}_{v})       \\
        \label{derivative of generalized momentum beta}
        &= \bm{\tau}_{v} + \bm{\tau}_{e} +\mathbf{\beta}(\mathbf{q}_{v}, \dot{\mathbf{q}}_{v}), 
\end{align}
where $\mathbf{\beta}(\mathbf{q}_{v}, \dot{\mathbf{q}}_{v}) =  {\mathbf{C}}^{T}(\mathbf{q}_{v}, \dot{\mathbf{q}}_{v})\dot{\mathbf{q}}_{v}-{\mathbf{g}}(\mathbf{q}_{v}) + \bm{\tau}_{f}$ is defined for the simplification.
To estimate external torque, a residual vector $\mathbf{r}\in \mathbb{R}^{n+6}$ and its dynamics are defined as follows,
\begin{align}
        \label{derivative of residue}
        \dot{\mathbf{r}} &= \mathbf{K}_0(\dot{\mathbf{p}} - \dot{\hat{\mathbf{p}}}) 
        \\
        \label{derivative of estimated momentum}
        {\dot{\hat{\mathbf{p}}}} &= \bm{\tau}_{v}+\mathbf{\beta}(\mathbf{q}_{v},\dot{\mathbf{q}}_{v})+\mathbf{r},
\end{align}
where $\mathbf{K}_0 = diag\{k_{0,i}\}>0$ is the positive diagonal gain matrix and $\dot{\hat{\mathbf{p}}}$ is the derivative of the estimated momentum. 
Integrating (\ref{derivative of residue}) results in 
\begin{equation}
\label{residue}
    \mathbf{r} = \mathbf{K}_0\left\{\mathbf{p}(t) - \mathbf{p}(0) - \int_{0}^{t}(\bm{\tau}_{v}+\mathbf{\beta}(\mathbf{q}_{v}, \dot{\mathbf{q}}_{v})+\mathbf{r})dt \right\}.
\end{equation}

This momentum observer can estimate the external torque using only $\mathbf{q}_{v}$, $\dot{\mathbf{q}}_{v}$, and $\bm{\tau}_{v}$. However, if the friction torque is dominant in the joint torque such as the robot with high gear reduction ratio and JTS is not available, accurate friction torque $\bm{\tau}_{f,j}$ should be estimated to calculate the residual in (\ref{residue}). Furthermore, the error of the dynamic model affects the precision of the external torque estimation.

\section{Proprioceptive External Torque Learning}
\label{section/Proposed Method}
In this work, the goal is to estimate the external joint torques of the floating base robot using only proprioceptive sensors. Especially, this study focuses on the external joint torque of the legs of the humanoid for walking control, but the proposed method is not restricted to a specific robot or limb. In the following sections, details on network architecture, data collection, and training will be introduced. Also, contact wrench reconstruction and contact wrench calibration for the inertia modification of the foot link are addressed.

\subsection{Network Architecture}
\label{Subsection/Proposed Method/Network Architecture}
Before specifying the neural network model for external torque estimation, proper input features for the learning should be defined first. From the analysis in Section \ref{Subsection/Preliminaries/External Torque Estimation using Proprioceptive Sensors}, the external torque of the robot is a function of generalized position, velocity, acceleration, and motor torque including the states of the base link. Specifically, $\mathbf{R}_{fb}$, $\mathbf{v}_{fb}$, $\bm{\omega}_{fb}$, $\dot{\mathbf{v}}_{fb}$, $\dot{\mathbf{\omega}}_{fb}$, $\mathbf{q}$, $\dot{\mathbf{q}}$, $\ddot{\mathbf{q}}$, and $\bm{\tau}_{m}$ are required for the estimation of the entire external torque of the robot. However, using all the states of the robot as an input of the network results in a large network size and redundant input features which can deteriorate the training result. Also, this work only focuses on estimating the external torques of the legs. Thus, a single network is designed to learn the external torque of a single leg separately by utilizing the sparsity of the dynamics induced by the kinematic tree structure of the humanoid \cite{featherstone2014rigid}. For example, the external torque of the left leg can be estimated using the full states of the floating base and the joint position, velocity, and acceleration of the left leg. As a result, the external torque of the left leg, $\bm{\tau}_{e,leg}\in \mathbb{R}^{l}$, can be expressed below
\begin{equation}
\label{equation/external torque of the left leg}
\begin{aligned}
    &\bm{\tau}_{e,leg} = f(\mathbf{q}_{leg}, \dot{\mathbf{q}}_{leg}, \ddot{\mathbf{q}}_{leg}, \bm{\tau}_{m,leg}, \mathbf{R}_{fb}, \mathbf{v}_{fb}, \bm{\omega}_{fb}, \dot{\mathbf{v}}_{fb}, \dot{\mathbf{\omega}}_{fb})
\end{aligned}
\end{equation}
where $\mathbf{q}_{leg}$, $\dot{\mathbf{q}}_{leg}$, and $\ddot{\mathbf{q}}_{leg} \in \mathbb{R}^{l}$ are the joint position, velocity, and acceleration of the left leg, respectively. $l$ is the number of joints in the left leg (6 for the humanoid robot used in this work). $\bm{\tau}_{m,leg}\in \mathbb{R}^{l}$ is the motor torque of the left leg joints. Note that, in the rest of this paper, external torque learning of the left leg is mainly addressed for simplicity, but the proposed method is also applied to the right leg. 

Despite the relationship in (\ref{equation/external torque of the left leg}), using the joint acceleration is not desirable because it is highly noisy. Furthermore, the orientation, the linear velocity, and the angular acceleration of the floating base (pelvis link) should be estimated because IMU typically measures linear acceleration and angular velocity, and only some high-performance IMU provides the estimated orientation and linear velocity optionally. In our previous work \cite{lim2021momentum} and in other works \cite{hwangbo2019learning, tran2020deep}, instead of estimating these unavailable input states explicitly, the time sequence of the available variable is used as the input of the network by assuming that the neural network can infer the unavailable variables intrinsically from the time sequence of the available variables. In this work, similarly, a neural network is trained to estimate the external torque directly only with the sequence of the available sensor information under several assumptions as below:
 \begin{itemize}
     \item Joint acceleration can be estimated with the time sequence of joint velocity.
     \item Orientation, angular acceleration, and linear velocity of the pelvis can be estimated with the time sequence of pelvis angular velocity, pelvis linear acceleration, joint position, and joint velocity \cite{rotella2014state}.
     \item Motor torque calculated from the current sensor $\bm{\tau}_{m,leg}$ has a reasonably small error when compared to the desired torque $\bm{\tau}_{d,leg}$ because of the high control accuracy of the motor current controller \cite{jung2017analysis}.
 \end{itemize}

Based on these assumptions, input variables and their sequence are defined as follows:
\begin{align}
    \label{equation/sequence of input}
    \mathbf{x}_{seq} &= [\mathbf{x}(k-h+1), \ ... \ , \ \mathbf{x}(k-1), \ \mathbf{x}(k)], \\
    \label{equation/input features of PETER}
    \mathbf{x}(k) &= [\mathbf{q}_{leg}(k),\ \dot{\mathbf{q}}_{leg}(k),\ \bm{\tau}_{d,leg}(k),\ {}^{p}\bm{\omega}_{fb}(k),\ {}^{p}\dot{\mathbf{v}}_{fb}(k)]
\end{align}
where $k$ is the current discrete control time, and $h$ is the length of the data sequence horizon. ${}^{p}\bm{\omega}_{fb}(k)$ and ${}^{p}\dot{\mathbf{v}}_{fb}(k)$ are the pelvis angular velocity and the pelvis linear acceleration in the pelvis local frame. It is notable that the required sensors for the input variables are fundamental and commonly available in most humanoid robots \cite{saeedvand2019comprehensive}. 

Then, the network, which estimates the external torque using the selected input sequence, can be formulated as $\hat{\bm{\tau}}_{e,leg}^T = f_{\theta}(\mathbf{x}_{seq})$.
$\hat{\bm{\tau}}_{e,leg} \in \mathbb{R}^{6}$ is the estimated external torque on the left leg joints, $f_{\theta}$ is the general expression of the neural network, and $\theta$ is the trainable network parameters. The reason for estimating the external torque instead of the external wrench directly is that the external torque has more general information than the external wrench on the specific link. Although the estimated external torque is used only for the contact wrench control in this paper, the estimated external torque also can be used for collision detection or impedance control on the other links, for example.

\begin{figure}[t]
\centering
\vspace*{0.0cm}
\includegraphics[width=0.9\linewidth]{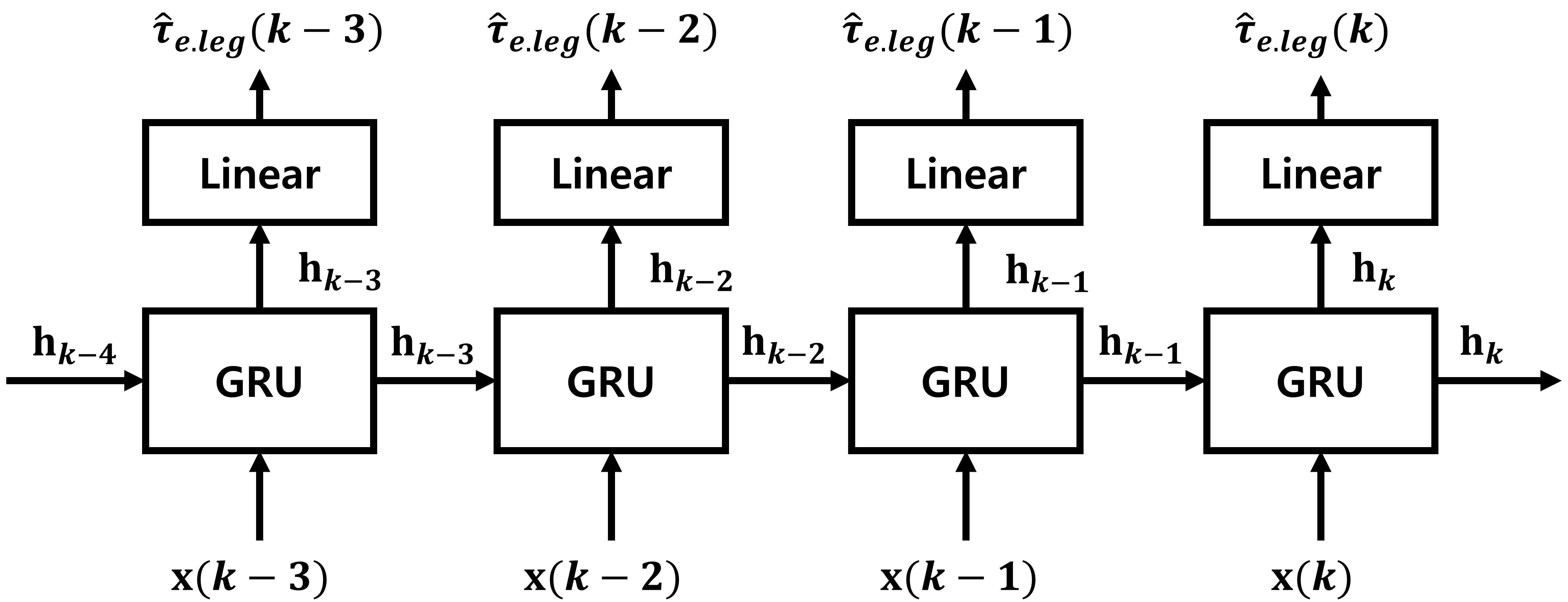}
\caption{The unfolded graph of GRU and a linear output layer. This GRU network takes the two vectors; current sensor data $\mathbf{x}(k)$, a hidden state vector $\mathbf{h}_{k-1}$ from the previous time step, and outputs $\mathbf{h}_{k}$. The linear output layer calculates the estimated external torque $\bm{\tau}_{e,leg}(k)$ from $\mathbf{h}_{k}$.}
\label{fig/GRU_Structure}
\end{figure}

\begin{table}[]
\centering
\caption{Summary of the GRU network structure and hyperparameters for training}
\label{table/Summary of the GRU network structure and hyperparameters for training}
\resizebox{\columnwidth}{!}{%
\begin{tabular}{@{}lclc@{}}
\toprule
\textbf{Network design} & \textbf{Value} & \textbf{Hyperparameter} & \textbf{Decision} \\ \midrule
Size of hidden state    & 150            & Initial learning rate   & 0.0001          \\
Size of input vector    & 24             & Batch size              & 64                \\
Size of output vector   & 6              & Epochs                  & 200               \\
Number of GRU layer     & 1              & Drop out                & 0                 \\
Frequency of input      & 100\,Hz          & Optimizer               & Adam(0.9, 0.999)  \\ \bottomrule
\end{tabular}%
}
\end{table}

\begin{figure*}[t]
\centering
\vspace*{0.0cm}
\includegraphics[width=0.95\linewidth]{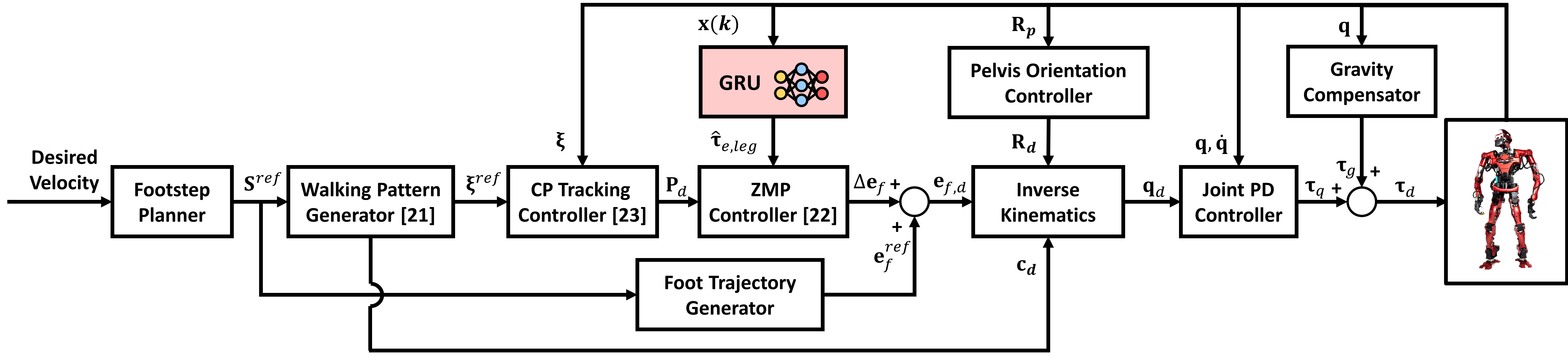}
\caption{Overall walking control framework and the estimated external torque feedback of the proposed GRU network to the ZMP controller. $\mathbf{S}^{ref}$ is the reference footsteps. $\bm{\xi}^{ref}$ is the reference CP trajectory. $\mathbf{P}_{d}$ is the desired ZMP. $\Delta\mathbf{e}_{f}$, $\mathbf{e}_{f}^{ref}$, and $\mathbf{e}_{f,d}$ are the modified, reference, and desired transformation of both legs, respectively. $\mathbf{R}_p$ and $\mathbf{R}_d$ are the current and desired rotation of the pelvis link, respectively. $\mathbf{c}_{d}$ is the desired center of mass position. $\mathbf{q}_{d}$ is the desired joint position. $\bm{\tau}_{q}$, $\bm{\tau}_{g}$, and $\bm{\tau}_{d}$ are the joint PD control torque, gravity compensation torque, and the desired joint torque, respectively.}
\label{fig/walking control framwork}
\end{figure*}

Among various neural network structures, RNN is specialized for training time-sequence data. Thus, Gated Recurrent Unit (GRU), a kind of RNN, is chosen as the structure of the network because GRU shows the best results in our training and validation data among several candidate networks [vanilla RNN, GRU, LSTM, TCN \cite{bai2018empirical}, MLP] with the similar number of network parameters and the same grid search of hyperparameter tuning. In contrast to our expectation that LSTM would perform better, it is observed that GRU outperformed LSTM slightly within our search space. We conjecture that the problem of external torque estimation requires only a short sequence of input history and that the long-term memory mechanism of LSTM embedded in the cell states may not contribute to better performance in this particular case. Some information on the GRU network design is summarized in TABLE \ref{table/Summary of the GRU network structure and hyperparameters for training}. As shown in Fig. \ref{fig/GRU_Structure}, the hidden state of the GRU is sent to the single linear layer to reduce the number of outputs to the dimension of the estimated external torque.

\subsection{Data Collection}
\label{Subsection/Proposed Method/Data Collection}
In this section, data collection procedures including the used sensors and data production method are introduced. Three kinds of sensors (encoder, IMU, and FTS) are used for the data collection. The encoder measures the joint angle $\mathbf{q}_{leg}$ and calculates the joint velocity $\dot{\mathbf{q}}_{leg}$ with numerical derivatives. 
An IMU consists of an accelerometer and a gyro sensor which measure a linear acceleration ${}^{p}\dot{\mathbf{v}}_{fb}$ and an angular velocity ${}^{p}\bm{\omega}_{fb}$ in the local coordinate, respectively. Lastly, an FTS is used to calculate the contact wrench under the foot of the left leg, $\mathbf{F}_{e,leg} \in \mathbb{R}^{6}$, from the raw FTS measurement, ${}^{FT}\mathbf{F}_{FT,leg} \in \mathbb{R}^{6}$, by calibrating the gravity of the foot link because the FTS of the leg is mounted between the ankle link and the foot link. However, the Coriolis \& centrifugal force and the inertia force of the foot link are ignored because it is relatively smaller than the gravity force during the walking experiments, and measurement of the foot link acceleration is highly noisy resulting in larger errors in the calculation of the external wrench. Accordingly, $\mathbf{F}_{e,leg}$ is calculated as
\begin{equation}
\begin{aligned}
\label{equation/external wrench calculation}
    {}^{c}\mathbf{F}_{e,leg} &= \ce{^{\emph{c}}_{\emph{FT}}\mathbf{X}}({}^{FT}\mathbf{F}_{FT,leg} - \ce{^{\emph{FT}}_{\emph{f}}\mathbf{X}}{}^{f}\mathbf{G}_{f}). \\
    {}_{B}^{A}\mathbf{X} &= 
    \begin{bmatrix}
        {}_{B}^{A}\mathbf{R} &\mathbf{0}_{3\times3}  \\
        -{}_{B}^{A}\mathbf{R} {}^{B}\mathbf{r}_{BA}\times & {}_{B}^{A}\mathbf{R} 
    \end{bmatrix}
\end{aligned}
\end{equation}
where ${}^{f}\mathbf{G}_{f} \in \mathbb{R}^{6}$ is the gravity wrench of the foot link in the local frame. ${}_{B}^{A}\mathbf{X}$ is the coordinate transformation from the coordinate frame \emph{B} to the coordinate frame \emph{A}, and ${}^{B}\mathbf{r}_{BA} \in \mathbb{R}^{3}$ is the vector expressing $\overrightarrow{BA}$ in \emph{B} coordinates. $\ce{^{\emph{c}}_{\emph{FT}}\mathbf{X}}$ is the transformation from the FT sensor position to the contact point on the foot (center of the sole) and $\ce{^{\emph{FT}}_{\emph{f}}\mathbf{X}}$ is the transformation from the center of mass of the foot link to the FT sensor position.

After that, the contact wrench is mapped to the external joint torque using the contact Jacobian ${}^{c}\mathbf{J}_{c} \in \mathbb{R}^{6\times6} $ which is from the pelvis link to the contact point of the left foot expressed in the contact foot local coordinates. The target value of the external torque learning can be obtained as 
\begin{equation}
\label{equation/external wrench mapping through jacobian transpose}
    \bm{\tau}_{e,leg} = {}^{c}\mathbf{J}_{c}^{T}{}^{c}\mathbf{F}_{e,leg}.
\end{equation}
Note that FTS is only used during the data collection procedure to obtain the target value for the network training ($\bm{\tau}_{e,leg}$), and FTS is no longer required for the inference of the external torque after the training.

Data is collected using the implemented walking controller in our robot \cite{herdt2010online, kajita2010biped, englsberger2016combining} by manually commanding the desired velocity of the robot randomly on the flat ground. The overall walking control framework for capture point (CP), $\bm{\xi}$, tracking is shown in Fig. \ref{fig/walking control framwork}.
The step length, turning angle, step period, and swing foot height are varied around the normally used values to collect the large distributed walking data. The step length varies in the closed interval [-0.10\,m, 0.15\,m] and the turning angle varies in the closed interval [-20$\degree$, 20$\degree$]. The step period is selected among four values \{0.6, 0.7, 0.9, 1.1\}\,s, and the maximum foot height is selected among three values \{4.0, 5.5, 7.0\}\,cm.
Additionally, external forces are intentionally applied to the swing foot and upper body during data collection in order to break a strong correlation between external torque and foot height. Otherwise, the network may rely on the swing foot position to infer the external torque.

The data is collected for 3 hours with a 100\,Hz sampling frequency (more than one million samples), and it is divided into the training data and validation data with a 9:1 ratio.

\subsection{Training}
\label{Subsection/Proposed Method/Training}
The network is trained using a typical deep-learning method of RNN. The parameters of the network are updated using the Truncated Back Propagation Through Time (TBPTT) method with 1\,s time horizon ($h=100$). All the hyperparameters for the training are summarized in TABLE \ref{table/Summary of the GRU network structure and hyperparameters for training}. For the parameter update, the Adam optimizer is used with the default betas (0.9, 0.999). The learning rate is scheduled to decrease linearly from the initial value in TABLE \ref{table/Summary of the GRU network structure and hyperparameters for training} to the 0.01 times smaller value during the first half of the epochs, which enables more stable and better learning results. It takes approximately 3 hours for the 200 epochs training using the Pytorch library and using a single desktop computer that has a single CPU (Intel Core i9-12900K × 24), and a single GPU (NVIDIA GeForce RTX 3090). However, it takes less than 30 microseconds to calculate the single network inference in the embedded robot computer (Intel Core i7-10700K × 8) by implementing the network in C++.

The loss function $L$ is the mean squared error between the target external torque value and the estimated torque formulated by 
\begin{equation}
\label{equation/loss function}
    L = \frac{1}{m}\sum_{i=1}^{m}  {(\hat{\bm{\tau}}_{e,leg,i} - \bm{\tau}_{e,leg,i})^{2}}
\end{equation}

\subsection{Contact Wrench Reconstruction and Calibration for Removal of FTS}
\label{Subsection/Proposed Method/Contact Wrench Reconstruction}
From the estimated external joint torque of the trained network $\hat{\bm{\tau}}_{e,leg}$, the external wrench on the foot $\hat{\mathbf{F}}_{e,leg}$ can be reconstructed again by the inverse of the $\mathbf{J}_{c}$. To enable numerically stable calculation near singularities, the damped pseudo-inverse is used as
\begin{equation}
\label{equation/external wrench reconstruction}
\hat{\mathbf{F}}_{e,leg} = \{\mathbf{J}_{c}^T\}^{+} \hat{\bm{\tau}}_{e,leg}. 
\end{equation}
Using the estimated contact wrench under the foot link, $\hat{\mathbf{F}}_{e,leg}$, the contact wrench feedback control can be realized for the stable locomotion of the humanoid.

However, to fully exploit the advantage of the proprioceptive external torque estimation, the real hardware sensor, FTS, should be replaceable with the neural network. Nonetheless, if the FTSs on the foot of the humanoid are removed, the inertia and size of the foot link which is included in the trained system will change. Therefore, it is necessary to deal with the removal of the FTS in kinematics and dynamics. Firstly, the contact position and the contact Jacobian matrix should be adapted according to the modified size of the foot link. Secondly, the modified gravity of the foot link should be reflected on the contact wrench estimation in (\ref{equation/external wrench reconstruction}). Otherwise, it is included in the estimated external wrench resulting in an estimation error. Accordingly, the external wrench can be calibrated for the FTS removal as,
\begin{equation}
\label{equation/external wrench reconstruction for foot link change}
{}^{c*}\hat{\mathbf{F}}_{e,leg}^{*} = \{{}^{c*}\mathbf{J}_{c*}^T\}^{+} \hat{\bm{\tau}}_{e,leg} + \ce{^{\emph{c*}}_{\emph{f}}\mathbf{X}}{}^{f}\mathbf{G}_{f} - \ce{^{\emph{c*}}_{\emph{f*}}\mathbf{X}}{}^{f*}\mathbf{G}_{f*}.
\end{equation}
where ${}^{c*}\hat{\mathbf{F}}_{e,leg}^{*}\in \mathbb{R}^{6}$ is the calibrated contact wrench estimation. $c*$ and $f*$ are the new contact position and new center of the mass of the modified foot link, respectively. ${}^{f*}\mathbf{G}_{f*}\in \mathbb{R}^{6}$ is the gravity force of the new foot link. As same as in (\ref{equation/external wrench calculation}), only the gravity force is considered in (\ref{equation/external wrench reconstruction for foot link change}).

\section{Experiments}
\label{Section/Experiments}
In this section, the learned external torque for the humanoid's leg is evaluated through real robot walking experiments. The estimated external torque by the proposed GRU network is validated on the test set, and it is used for the ZMP control of the supporting leg in Section \ref{Subsection/Experiments/Test Results and Contact Force Admittance Control for Humanoid Locomotion}. 
Then, when the inertia of the robot is modified, the consistent performance of the external wrench estimation for humanoid walking control is shown in Section \ref{Subsection/Experiments/General ZMP Control Performance for Hardware Modification of the Robot}.

Through all the experiments, a torque-controllable full-size humanoid robot, TOCABI, is used, and only the results of the left leg are demonstrated because the right leg has similar results to the left leg. For the robot's hardware specifications, TOCABI has 33 DoF (16 in both arms, 12 in both legs, 3 in the waist, and 2 in the neck), the height is approximately 1.8\,m, and the weight is around 100\,kg. The robot has an FTS (ATI Mini85 SI-1900-80) on each foot, an IMU sensor (Lord Microstrain 3DM-GX5-25) on the pelvis, and a motor side encoder in each joint. The torque control frequency is 2\,kHz. Note that the positive direction of the x-axis is the forward direction of the robot, and the positive direction of the z-axis is the vertically upward direction aligned with gravity.

\begin{figure*}[!ht]
\centering
\subfloat[]{\includegraphics[width=1.0\linewidth]{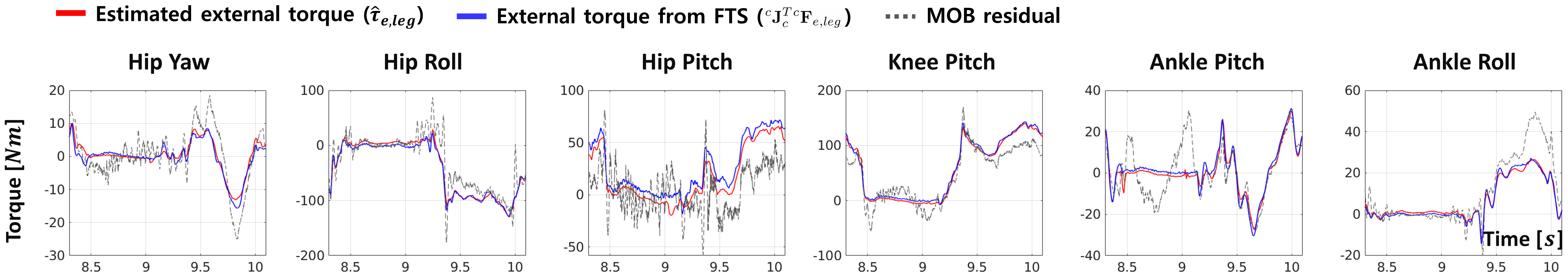} \label{fig/estimated external torque}}

\subfloat[]{\includegraphics[width=1.0\linewidth]{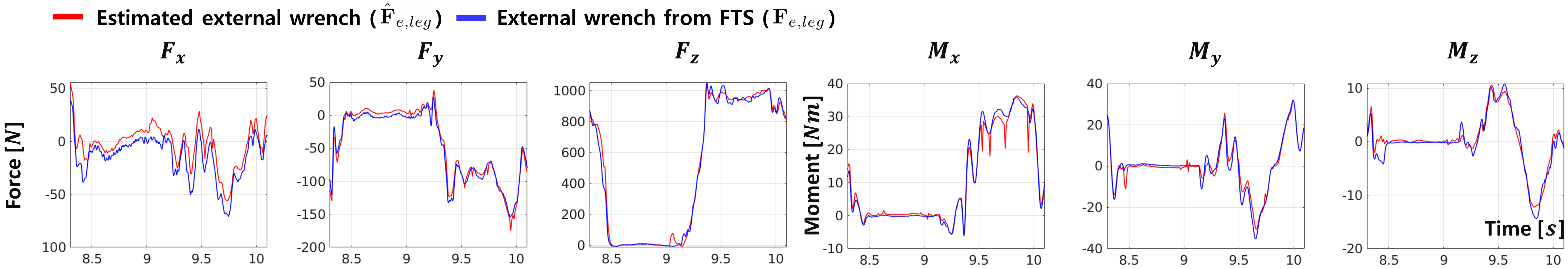} \label{fig/estimated external wrench}}

\caption{(a) The estimated external joint torque of the left leg by the proposed GRU network (red), the measured external torque by FTS (blue), and the estimated external torque by MOB with friction model (gray). (b) The estimated contact wrench by projecting the estimated external torque from the network (red), and the measured contact wrench by FTS (blue).}
\label{fig/estimated external torque wrench}
\end{figure*}

\subsection{Test Results and Contact Wrench Feedback Control for Humanoid Locomotion}
\label{Subsection/Experiments/Test Results and Contact Force Admittance Control for Humanoid Locomotion}
The trained GRU network is evaluated on the real robot for the estimation of the external torque and the contact wrench while the robot is walking forward. The robot walked 1.0\,m straight forward with 0.1\,m step length, 0.055\,m maximum step height, and 0.9\,s step period (0.6\,s for single support, 0.3\,s for double support) for each step. The estimated external torque and the target value calculated by the FTS for the left leg joint are displayed during one walking cycle [8.3\,s, 10.1\,s] in Fig. \ref{fig/estimated external torque wrench}(a). For the first half of the time [8.45\,s, 9.05\,s], the left leg is in the swing leg phase and it transits to the double support phase after 9.05\,s. And, after 9.35\,s, the left leg becomes the supporting leg. A model-based external torque estimation method, MOB with friction model \cite{lee2015sensorless}, is also implemented for the comparison because MOB requires the same proprioceptive sensors as the proposed method. For the implementation of MOB to the floating-based robot, the estimated linear velocity of the pelvis link is used which is estimated by the complementary filter using the joint states and IMU measurements. 

As shown in Fig. \ref{fig/estimated external torque wrench}(a), all the estimated torques (red) track the target external torque values (blue) closer than the estimated disturbances by the MOB (dotted gray). Even though the friction torque is reflected in the MOB, the estimation error is still large because of the robot modeling errors, and the other ignored disturbances, for example, joint stiffness and load-dependent friction.

Mean absolute error (MAE), root mean square error (RMSE), and relative MAE (rMAE) of the estimated torque for each joint are shown in TABLE \ref{table/test results estimated external torque}. rMAE refers to the relative percentage of the MAE over the maximum external torque measured by the FTS. According to the results in the last low of TABLE \ref{table/test results estimated external torque}, MAEs and RMSEs of the MOB residual are approximately 3 times larger than the ones of the GRU except for the fourth joint (5 times larger). The estimated external torque by the proposed learning method can be used for the ZMP control to balance the robot while walking. However, the robot failed to maintain balance using the MOB residual for the contact wrench feedback control of the supporting leg because of its large estimation errors.

From the estimated external torque of the trained network $\hat{\bm{\tau}}_{e,leg}$, the external wrench on the foot $\hat{\mathbf{F}}_{e,leg}$ can be calculated as in (\ref{equation/external wrench reconstruction}).
The estimated external wrench by the network and the measured external wrench by FTS on the left foot are shown in Fig. \ref{fig/estimated external torque wrench}(b). Also, the statistics of the external wrench estimation errors are summarized in TABLE \ref{table/estimated external wrench error}. The MAE in $F_z$ is 19.743\,N which is 1.7$\%$ of the maximum force on the z-axis while the rMAE of $F_x$ is 11.3$\%$. However, $F_z$, $M_x$, and $M_y$ are only used to calculate the ZMP of the robot.

\begin{table}[!t]
\centering
\caption{Statistics of the external torque estimation errors on the real robot straight walking data with GRU and MOB}
\label{table/test results estimated external torque}
\resizebox{0.896\columnwidth}{!}{%
\begin{tabular}{@{}cccccccc@{}}
\toprule
\multirow{2}{*}{\textbf{Estimation}} & \multirow{2}{*}{\textbf{Statistics}} & \multicolumn{6}{c}{\textbf{Joint \#}}                                       \\ \cmidrule(l){3-8} 
                                     &                                      & \textbf{1} & \textbf{2} & \textbf{3} & \textbf{4} & \textbf{5} & \textbf{6} \\ \midrule
\multirow{3}{*}{GRU}                 & MAE (Nm)                             & \cellcolor[HTML]{C0C0C0}1.043      & \cellcolor[HTML]{C0C0C0}3.796      & \cellcolor[HTML]{C0C0C0}7.286      & \cellcolor[HTML]{C0C0C0}3.741      & \cellcolor[HTML]{C0C0C0}1.795      & \cellcolor[HTML]{C0C0C0}1.091      \\
                                     & RMSE (Nm)                            & 1.354      & 4.826      & 8.161      & 4.517      & 2.276      & 1.406      \\
                                     & rMAE (\%)                            & 6.3        & 2.7        & 8.5        & 2.4        & 4.2        & 2.9        \\ \midrule
\multirow{3}{*}{MOB}                 & MAE (Nm)                             & 2.845      & 12.830     & 23.597     & 18.801     & 6.115      & 4.030      \\
                                     & RMSE (Nm)                            & 3.705      & 19.686     & 26.762     & 22.425     & 8.165      & 6.860      \\
                                     & rMAE (\%)                            & 17.1       & 9.2        & 27.6       & 11.8       & 14.4       & 10.6       \\ \midrule
MOB/GRU                              & MAE/MAE                              & 2.728      & 3.380      & 3.239      & 5.026      & 3.407      & 3.694      \\ \bottomrule
\end{tabular}%
}
\end{table}

\begin{table}[!t]
\centering
\caption{Statistics of the external contact wrench estimation errors on the real robot straight walking data with GRU}
\label{table/estimated external wrench error}
\resizebox{0.896\columnwidth}{!}{%
\begin{tabular}{@{}ccccccc@{}}
\toprule
\multirow{2}{*}{\textbf{Statistics}} & \multicolumn{6}{c}{\textbf{Wrench}}                             \\ \cmidrule(l){2-7} 
                            & $F_x$ (N) & $F_y$ (N) & $F_z$ (N) & $M_x$ (Nm) & $M_y$ (Nm) & $M_z$ (Nm) \\ \midrule
MAE                         & 10.329 & 5.938  & 19.743 & 1.191   & 1.717   & 0.831   \\
RMSE                        & 12.232 & 7.412  & 31.139 & 1.839   & 2.420   & 1.113   \\
rMAE (\%)                   & 11.3   & 3.5    & 1.7    & 3.1     & 3.7     & 5.3     \\ \bottomrule
\end{tabular}%
}
\end{table}

\begin{figure}[ht]
\centering
\vspace*{0.0cm}
\includegraphics[width=0.99\linewidth]{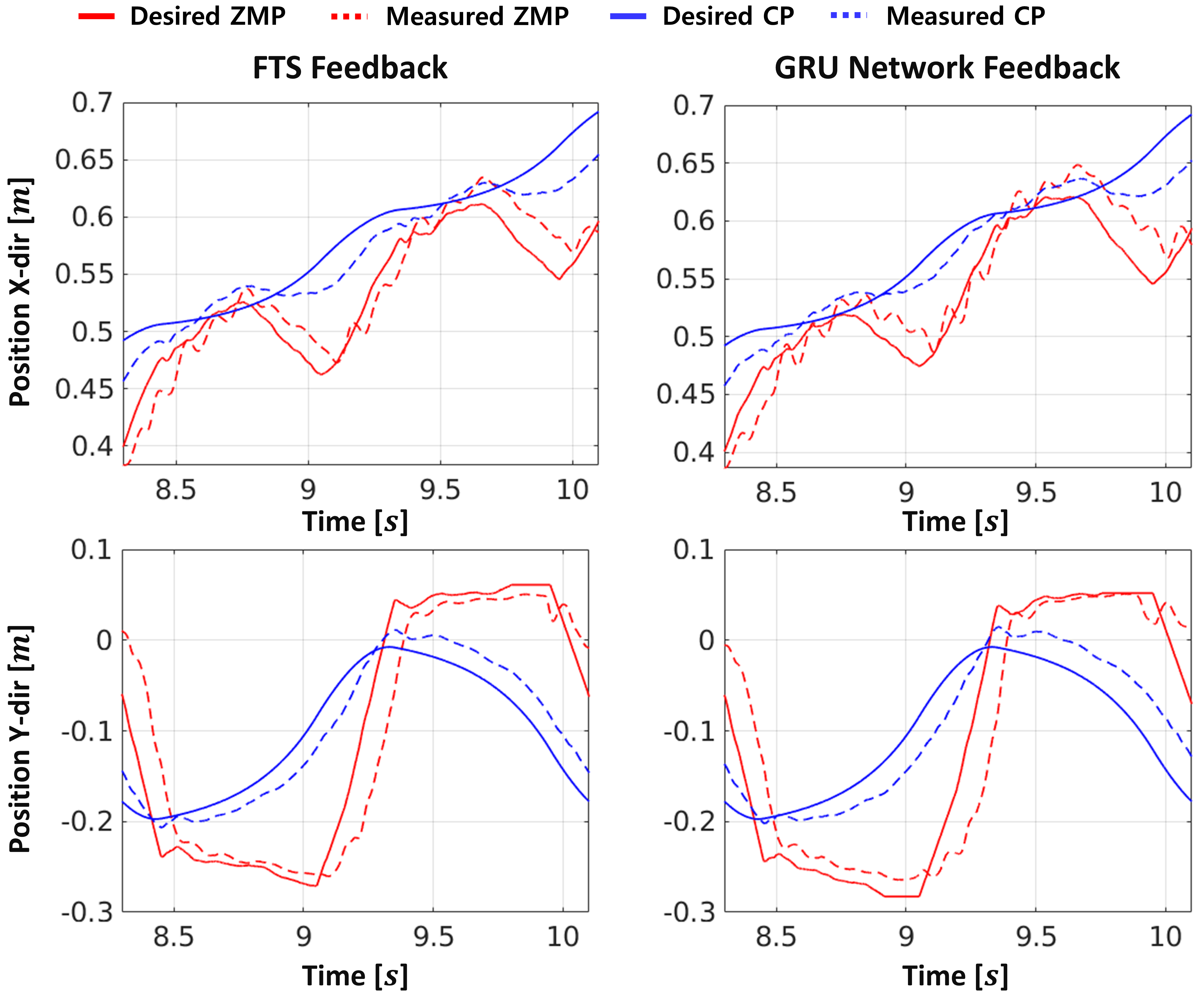}
\caption{ZMP and CP trajectories when the humanoid is walking forward according to the contact wrench feedback methods.}
\label{fig/ZMP CP trajectory}
\end{figure}

\begin{figure}[ht]
\centering
\vspace*{0.0cm}
\includegraphics[width=1.00\linewidth]{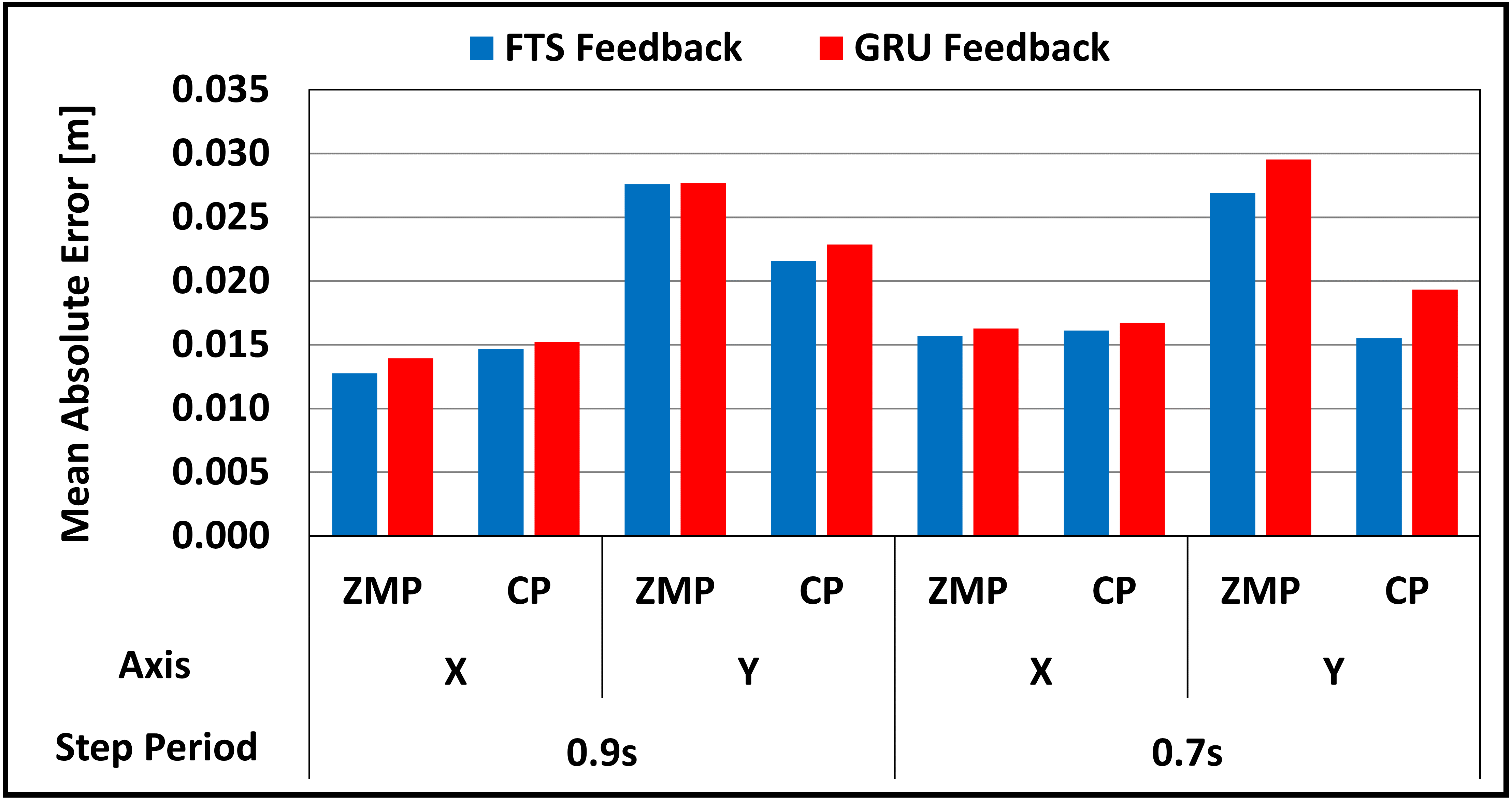}
\caption{Comparison of ZMP and CP tracking errors in x and y directions according to the contact force sensing method.}
\label{fig/ZMP CP errors in testing data}
\end{figure}

The estimated external wrench, $\hat{\mathbf{F}}_{e,leg}$, is used as the feedback signal for the ZMP control \cite{kajita2010biped} while the robot is walking forward with a step period of 0.9\,s. With the precise estimation performance of the network, the humanoid robot can walk robustly by substituting the FTS measurements with the network estimation, $\hat{\mathbf{F}}_{e,leg}$. The ZMP and CP trajectories in x and y directions according to the contact wrench feedback signal are demonstrated in Fig. \ref{fig/ZMP CP trajectory}. The two figures in the left column of Fig. \ref{fig/ZMP CP trajectory} are the ZMP (red) and CP (blue) trajectories when the FTS measurement is used. The right column shows the result of the walking using the GRU network estimations instead of FTS measurement. As shown in Fig. \ref{fig/ZMP CP trajectory}, it is difficult to distinguish between two walking trajectories with the FTS measurement and GRU estimation. 

The ZMP and CP tracking errors for the experiments with 0.9\,s and 0.7\,s step periods are displayed in Fig. \ref{fig/ZMP CP errors in testing data}. Although the ZMP and CP errors of the network feedback experiments (red) are higher than the errors of the FTS feedback (blue), the errors are still low enough for the robot to walk stably. Moreover, even for the step period of 0.7\,s, stable walking is realized with the network wrench estimation similar to the 0.9\,s step period experiment without any difficulties.

\begin{figure}[t]
\centering
\vspace*{0.0cm}
\includegraphics[width=0.90\linewidth]{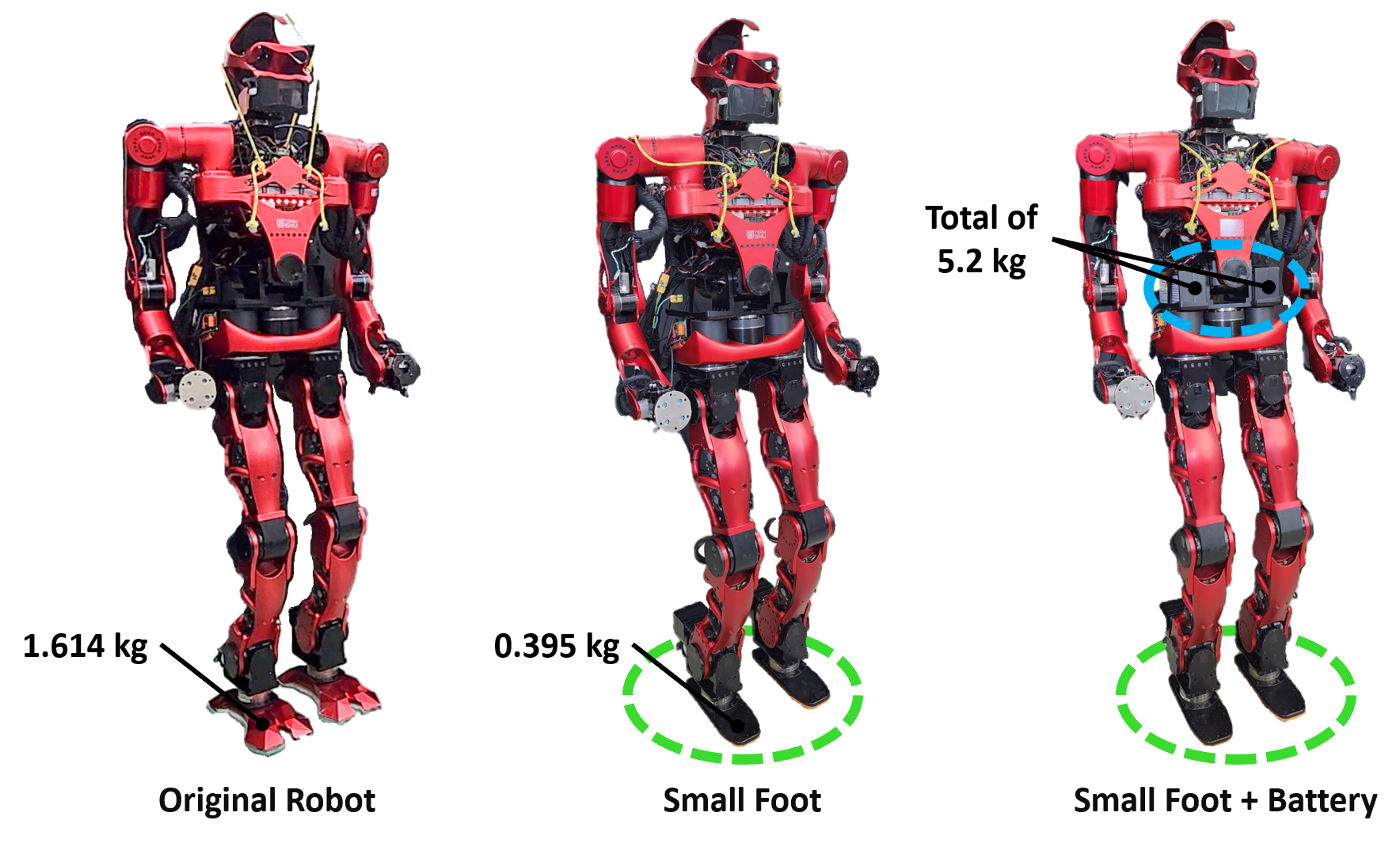}
\caption{The original robot hardware used for data collection and two different hardware settings (Small Foot and Small Foot + Battery) for the validation of the consistent performance of the proposed method with respect to the inertia changes.}
\label{fig/hardware modification hardware}
\end{figure}

\begin{figure}[t]
\centering
\vspace*{0.0cm}
\includegraphics[width=1.0\linewidth]{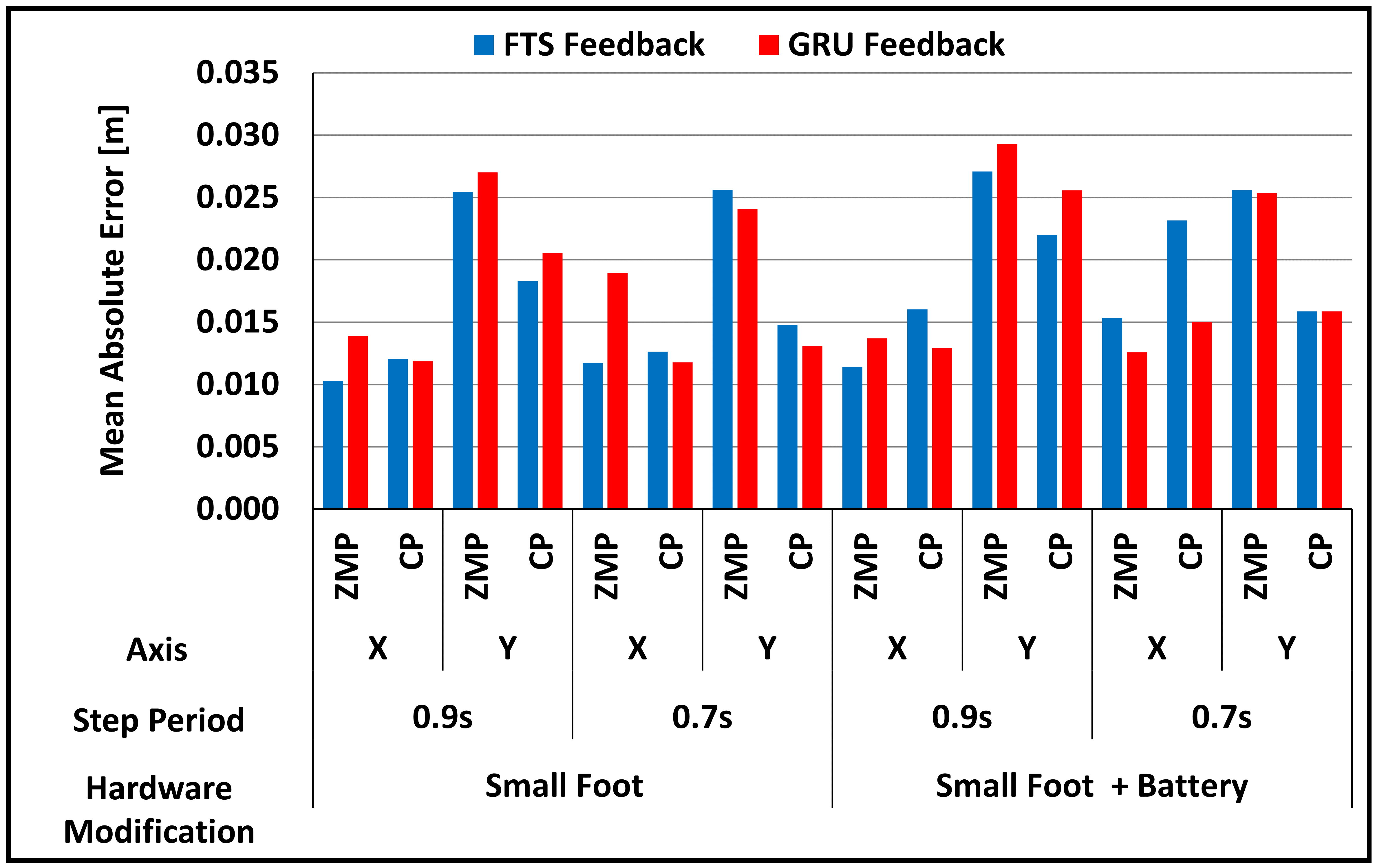}
\caption{MAE of ZMP and CP in each axis for two contact wrench sensing methods according to the hardware settings when the humanoid robot walks forward for 1.0\,m with step periods 0.9\,s and 0.7\,s. The blue bar represents the ZMP and CP errors when the robot uses FTS measurement for the ZMP feedback control. The red bar represents the errors when the robot uses GRU network estimations for the ZMP feedback control.}
\label{fig/ZMP CP MAE hardware modification}
\end{figure}

\subsection{Consistent Walking Control Performance for Modification of Robot Foot Link and Upper Body Mass}
\label{Subsection/Experiments/General ZMP Control Performance for Hardware Modification of the Robot}
In this experiment, the possibility of the FTS replacement with the neural network is validated and the estimation performance across different hardware settings is analyzed. 

However, instead of removing FTS in the robot, the experimental setting is designed to imitate the situation where the FTS is removed because the measured contact wrench is necessary for the calculation of the estimation error of the network and the calculation of the measured ZMP. As shown in Fig. \ref{fig/hardware modification hardware}, the robot is equipped with lighter and smaller foot links to mimic the absence of the FTS. 
The original foot link weighs 1.614\,kg and the small foot weighs 0.395\,kg. The decreased mass of a single-foot link is over 1.2\,kg which is approximately 2 times larger than the mass of a single FTS (0.635\,kg). The size of the foot sole is also decreased from $0.30\,m \times 0.13\,m$ (original foot) to $0.26\,m \times 0.10\,m$ (small foot).
For the small foot modification, the modified foot link which is included in the trained system is explicitly reflected in the contact wrench estimation as described in (\ref{equation/external wrench reconstruction for foot link change}). Furthermore, the model of the robot, the control gains in the admittance controller \cite{kajita2010biped}, and the desired ZMP constraints are adjusted for the small foot to realize stable walking control performance. 

Moreover, another hardware setting is also prepared as shown in the third figure of Fig. \ref{fig/hardware modification hardware} where two battery cells (total of 5.2\,kg, approximately 5\% of the robot mass) are added to the upper body of the robot. This experimental setting is designed to show the adaptation ability of the trained network when the mass of the upper body is increased compared to the original robot setting. Contrary to the small foot setting, battery cells increase the inertia outside of the trained system only resulting in the increased magnitude of the external wrench on the supporting foot. Thus, the estimated external wrench is not calibrated explicitly for the battery, and only the model of the robot is adjusted. Note that all the same procedures are also applied for the experiments with the FTS measurement when the link inertia is modified.

A total of eight experiments are conducted with all combinations between two different hardware settings (small foot and small foot + battery), two different step periods (0.9\,s and 0.7\,s), and two different contact wrench sensing methods (FTS and GRU). The experimental results are summarized in Fig. \ref{fig/ZMP CP MAE hardware modification} for the ZMP, and CP errors while the robot is walking forward 1.0\,m with the 0.1\,m step length. Through all cases, the humanoid robot could walk stably on flat ground with the estimated contact wrench feedback despite the inertia modifications.
Generally, walking experiments with a 0.9\,s step period show that the experiment with network wrench feedback (red) has larger errors in the ZMP and CP tracking than the experiment with FTS measurement (blue) as shown in Fig. \ref{fig/ZMP CP MAE hardware modification} while, in experiments with a 0.7\,s step period, the ZMP and CP errors with GRU feedback are lower than the errors with FTS feedback.
In spite of such differences in statistics of ZMP and CP errors, the behaviors of the robot are not distinguishable between the two force sensing methods (See the \href{https://www.youtube.com/watch?v=a2eLoIsEF9I}{supplementary video}). 

These experimental results validate the generalization ability of the proposed learning method aided by the model-based calibration when the robot's links are changed, especially for the humanoid locomotion task. 
When the size of the foot link is changed, it is difficult for the network to adapt to the kinematic changes without additional training. Therefore, the model-based calibration method introduced in \ref{Subsection/Proposed Method/Contact Wrench Reconstruction} is used not only for the inertia changes but also for the kinematic changes of the foot link. However, for the increased mass of the upper body by the batteries, the proposed network can estimate the external wrench without further training or model-based calibration.

The lesson from this experiment is that even though the distribution of the desired torque and external torque is shifted by the inertia changes, the network can estimate the external torque without losing generalization thanks to the distributed random walking data. Deep learning method shows superior generalization ability for interpolation problems, but it shows poor performance for extrapolation problems. Therefore, this consistent estimation performance can maintain as long as the input and output signals are in the range of the training data. Consequently, collecting widely distributed training data around the target task is crucial for this data-driven method.

\section{Conclusion}
\label{Section/Conclusion}
In this study, a novel external torque learning is proposed. From the analysis of the dynamics and under several assumptions, the GRU network architecture is designed. For the network training, largely distributed random walking data of the humanoid robot is collected. Finally, it is validated that the external torque can be estimated accurately only using the proprioceptive sensors (encoder and IMU) through real robot experiments. Especially, the estimated external torque from the network is utilized for the ZMP control of the humanoid robot. Moreover, although the inertia of the foot link and upper body changes, consistent performance of the contact wrench estimation is demonstrated by using the same network and using the model-based calibration through extensive locomotion experiments. 

We believe that the proposed learning method can reduce the production cost, inertia, hardware complexity, and the possibility of system failure by substituting the FTS hardware with the neural network, especially, when the humanoids are manufactured in large quantities.

For future works, improving the performance of the external torque estimation for various tasks will be studied by combining the model-based method and data-driven learning method. Furthermore, using the estimated external torque, safety-oriented studies to realize impact-aware and impact-reactive humanoids will be conducted.

\addtolength{\textheight}{0cm}   





\bibliographystyle{Bibliography/IEEEtran}
\bibliography{Bibliography/IROS2023}\ 
\end{document}